\pdfoutput=1
\documentclass{article} 
\usepackage{iclr2027_conference,times}

\usepackage{amsmath,amsfonts,bm}

\def\eqref#1{equation~\ref{#1}}

\def\1{\bm{1}}

\DeclareMathAlphabet{\mathsfit}{\encodingdefault}{\sfdefault}{m}{sl}
\SetMathAlphabet{\mathsfit}{bold}{\encodingdefault}{\sfdefault}{bx}{n}

\usepackage{hyperref}
\usepackage{url}

\usepackage[table]{xcolor}   
\usepackage{booktabs}
\usepackage{multirow}
\usepackage{array}
\usepackage{makecell}                 
\usepackage{graphicx}                 

\definecolor{headbg}{HTML}{DDE5EE}
\definecolor{groupbg}{HTML}{F2F6FA}

\newcommand{\grp}[1]{\multicolumn{10}{@{}>{\columncolor{gray!8}[0pt][0pt]}l@{}}{\textit{#1}}}

\definecolor{headergray}{gray}{0.90} 
\newcommand{\hd}[1]{%
  {\renewcommand{\arraystretch}{1}\begin{tabular}[t]{@{}c@{}}#1\end{tabular}}}
\newcommand{\first}[1]{\textbf{#1}}        
\newcommand{\second}[1]{\underline{#1}}    
\newcommand{\meth}[1]{\hspace{1em}#1}      
\newcommand{\ws}[1]{#1\,s}                 
 
\newcommand{\fitwidth}[1]{%
  \resizebox{\ifdim\width>\linewidth\linewidth\else\width\fi}{!}{#1}}

\definecolor{ablgreen}{HTML}{2E7D32} 
\newcommand{\abldn}[1]{\,{\scriptsize\textcolor{red}{($-$#1)}}}
\newcommand{\ablup}[1]{\,{\scriptsize\textcolor{ablgreen}{(+#1)}}}

\usepackage[capitalize]{cleveref}

\usepackage{xspace}
\newcommand{\ourmethod}{\texttt{VarMAE\xspace}}

\usepackage{arydshln}       

\newcommand{\winlabel}[1]{\smash{\raisebox{\dimexpr\arraystretch\ht\strutbox+2pt-0.7ex\relax}{#1\,s}}}

\newcommand{\ablzero}{\,{\scriptsize(0.000)}}

\usepackage{enumitem}

\title{
    Variable Input Masked Autoencoders for Wrist-Worn Accelerometry
}

\author{%
\textbf{Harish Haresamudram}$^{1}$\thanks{Corresponding author: \texttt{hkh@illinois.edu}}\quad
\textbf{Wanting Mao}$^{1}$\quad
\textbf{Aditya Radhakrishnan}$^{1}$ \\
\textbf{Maxwell A.~Xu}$^{1, 2}$\quad
\textbf{James M.~Rehg}$^{1}$ \\
\textnormal{$^{1}$University of Illinois Urbana-Champaign \quad $^{2}$Google}
}

\iclrfinalcopy 
\begin{document}

\maketitle
\lhead{Preprint. Under review.}
\fancyfoot[C]{}\fancyfoot[R]{\thepage}
\fancyfoot[L]{\footnotesize \textit{This paper is currently under review. Please contact \texttt{hkh@illinois.edu} if you have any questions.}}

\begin{abstract}
    Large-scale repositories of wearable sensor data have enabled the development of foundation models for human activities and behavior.
    However, existing models are pretrained using fixed input durations and sampling rates, despite substantial heterogeneity in recording conditions, downstream tasks, and deployment constraints. 
    We introduce \ourmethod{}, a variable input masked autoencoder that learns a single representation across sampling rates and temporal contexts. 
    During pretraining we construct two views of the same signal: a long duration high frequency view, and a short duration, lower frequency view, obtained via random cropping and downsampling.
    We perform masked reconstruction within each view and introduce a cross-resolution masked reconstruction objective that uses visible tokens from the long view to predict masked tokens from the short view. 
    Cropping and downsampling impose complementary information bottlenecks, encouraging the model to learn representations that remain effective across input durations and sampling rate changes.
    We pretrain \ourmethod{} on the large-scale NHANES dataset, producing a single pretrained encoder matching or outperforming controlled foundation model baselines on eight benchmark datasets.
    With only 3.82M parameters, \ourmethod{} is parameter efficient and shows that a single foundation model can achieve broad transfer capabilities across sensing configurations.
\end{abstract}


\section{Introduction}

The popularity and widespread use of wearables has led to transformative progress across a wide gamut of health applications, including sleep analysis \citep{yuan2024selfsleep}, monitoring Parkinson's disease \citep{del2016free}, fall detection \citep{chatzaki2016human}, and predicting flu \citep{merrill2023homekit2020}. 
Their ubiquitous nature allows biosignals to be collected in everyday settings, leading to vast amounts of sensor data \citep{schuetz2026openmhc}.
This in turn has motivated foundation model (FM) development, as unlabeled physiological data can be collected at scale whereas obtaining high quality annotations is both expensive and time consuming \citep{haresamudram2025past}.

Among wearable sensing modalities, accelerometry is widely used to capture human movements and underpins applications ranging from gesture recognition \citep{bonazzi2026openwatch} to analysis of daily activities and mobility \citep{yang2010review}.
However, accelerometer FMs face a fundamental challenge: \textit{there is no single gold-standard input configuration in wearable sensing.}
Downstream applications differ substantially in both the temporal context they require and the sampling rate at which the signal is sampled.
For example, the OpenWatch gesture benchmark \citep{bonazzi2026openwatch} utilizes one second windows recorded at 100 Hz to recognize short gestures such as pinches and slides. 
In contrast, \cite{yuan2024self} utilize ten second windows at 30 Hz to model daily living activities in naturalistic conditions. 
These differences stem from factors such as the timescale of the activity being observed, practical constraints such as the hardware platform, and other deployment requirements.

Existing accelerometer FMs largely sidestep this heterogeneity. 
As noted by \cite{tarale2026bio}, the window duration and sampling rate are typically fixed during pretraining and downstream evaluation occurs in similar input configurations.
While there are adaptations like resampling or padding that can be applied, they undermine a key promise of foundation models: a \textit{single pretrained model} that transfers broadly across many applications.
It remains unclear how well representations learned under one input configuration transfer to others.

Our approach, \ourmethod{}, addresses this limitation through a variable input masked autoencoder (MAE) trained using two views of the same sensor data: \textit{(i)} a long duration, high frequency view that captures the full context; 
and \textit{(ii)} a short duration, lower frequency view which is randomly cropped from the long view, allowing the model to adapt to different frequencies.
Both views are masked, and the masked regions in the short view are also masked in the long view to prevent information leakage about the masked target.
In addition to within-view masked reconstruction, we introduce \textit{cross-resolution masked reconstruction}, where the visible regions from the long view condition the prediction of the masked tokens of the short view. 

Our key idea is to employ \textit{cropping and downsampling as information bottlenecks.}
On one hand, downsampling removes fine-grained temporal information while preserving the underlying signal; on the other, cropping removes the broader context surrounding the movements. 
Cross-resolution masked reconstruction bridges these bottlenecks by using information from the long context, high rate view to predict masked tokens at the low rate cropped view. 
This setup \textit{exposes a single encoder to a range of sampling rates and window durations} during pre-training, enabling it to operate across a range of sensing configurations rather than specializing to a single one.

Architecturally, \ourmethod{} retains the standard MAE's Transformer encoder-decoder, but extends it with rate flexibility in both input and output.
We define patches over a fixed duration of time, causing the number of samples within each patch to vary based on the sampling rate.
For example, a patch duration of 0.2s is 10 samples of sensor data at 50Hz but two samples at 10Hz.
To accommodate this, a \textit{variable-rate tokenizer} maps these rate dependent patches to a single fixed dimension token vector before the Transformer encoder, and correspondingly, a \textit{variable-rate reconstruction head} reconstructs the decoder's outputs at the target sampling rate.

\ourmethod{} is pretrained on NHANES \citep{cdc2010nhanes, troiano2008physical}, which is a publicly available large-scale wrist accelerometer dataset comprising approx.\ one week of data from around 16k subjects. 
We evaluate our \textit{single pretrained encoder} on eight human activity recognition (HAR) datasets covering a wide range of activities, subject counts, recording conditions, and input configurations.
Despite using only 0.7M params, \ourmethod{} outperforms general purpose time-series FMs that are 30--550$\times$ larger.
Scaling our method to 3.82M params further improves performance, and the single encoder matches or outperforms the specialist FMs we pretrain for each input configuration on all datasets. 
These results demonstrate that a single accelerometer FM can effectively transfer across substantially different downstream sensing configurations.

Our contributions are threefold:
\begin{enumerate}[label=(\arabic*), leftmargin=1.6em, itemsep=1pt, topsep=1pt, parsep=0pt]
    \item We introduce a variable input temporal masked autoencoder that enables a single Transformer encoder-decoder to operate across different sampling rates and window sizes, rather than specializing to a fixed sensing configuration.
    \item We propose \textit{cross-resolution masked reconstruction} to jointly learn across temporal context and resolution by using a longer, high rate view to reconstruct masked information in a shorter, lower-rate view.
    This improves representation quality, particularly at low sampling rates.
    \item Using a \textit{single pretrained encoder}, we demonstrate broad and parameter-efficient transfer across eight benchmarks spanning 1--10\,s window sizes.
\end{enumerate}

\section{Related Work}

In wearable computing, the development of FMs has received substantial interest in recent years, enabled by the availability of large-scale datasets such as UK Biobank \citep{doherty2017large}, NHANES \citep{troiano2008physical}, and OpenMHC \citep{schuetz2026openmhc}. 
The goal is to develop generic models that can support a wide range of downstream applications, e.g., gait analysis \citep{kavanagh2008accelerometry, tao2012gait}, fall detection \citep{bagala2012evaluation}, and activity recognition \citep{ordonez2016deep}.
Moreover, self-supervised learning (SSL) has been shown to reduce reliance on annotated data \citep{haresamudram2022assessing}, and produce representations that are robust to class imbalance \citep{liu2021self}.

For accelerometry, an early SSL approach involved augmentation prediction \citep{saeed2019multi}.
\cite{yuan2024self} later scaled this approach to 700k person days of data from UK Biobank, and the resulting representations were used for tasks like activity recognition, analysis of the association between sleep and mortality \citep{yuan2024selfsleep}, and animal behavior recognition \citep{hoffman2024benchmark}. 
More generally, two dominant paradigms emerged: reconstruction and contrastive learning. 
Contrastive methods learn by training to reduce the distance between positive pairs while increasing it between negative pairs.
Such pairs have been constructed using specialized augmentations \citep{chen2020simple, tang2020exploring}, time-frequency consistency \citep{zhang2022self}, future timestep prediction \citep{haresamudram2021contrastive}, and learned similarity functions \citep{xu2024rebar, xu2025relcon}, among other strategies.
Reconstruction-based methods recover either the entire input \citep{haresamudram2019role} or intentionally masked portions of it, as in MAEs \citep{he2022masked}.
This setup has recently shown great effectiveness when scaled to large datasets and models \citep{narayanswamy2025scaling, xu2025lsm, tarale2026bio}.

A limitation shared by most existing SSL approaches is that each pretrained model fixes the sensing configuration, i.e., the window size and sampling rate, throughout training and evaluation.
For example, \cite{yuan2024self} downsample UK Biobank data to 30 Hz and pre-train/evaluate using either five- or ten-second windows.
Similarly, \cite{xu2025relcon} pre-train separate models for 2.56s and 2s windows in separate experiments. 
The input configurations used by prior FMs are summarized in the Appendix (\cref{tab:input_config_for_fms}).
Therefore, existing approaches largely address the differences in sensing configurations through configuration-specific pretraining.
In contrast, we train a \textit{single accelerometer FM} across sampling rates and window sizes. 
Our model operates directly on heterogeneous configurations, without requiring any adaptations or setting-specific pretraining. 

\section{Methodology} \label{sec:approach}
\begin{figure*}
    \centering
    \includegraphics[width=1\linewidth]{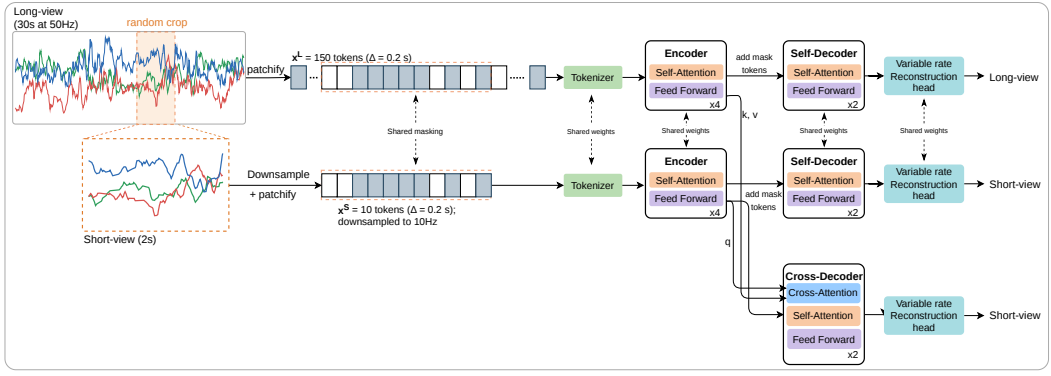}
    \caption{
    \ourmethod{} is a variable input MAE trained using two views: first, a long duration, high frequency view, and second, lower frequency, short duration view.
    In addition to within-view masked reconstruction, we also perform cross-resolution masked reconstruction.
    }
    \label{fig:approach_figure}
\end{figure*}

Our goal is to \textit{pre-train a single encoder} that can reliably transfer across a variety of window durations and sampling rates. 
To achieve this, \ourmethod{} has two key components: \textit{(i)} a variable-rate masked autoencoder that processes accelerometer signals at different sampling rates and window durations; and \textit{(ii)} a cross-resolution masked reconstruction objective where a short, downsampled view is reconstructed using context from a longer, higher frequency view of the same recording. 

\subsection{Variable-Rate Masked Autoencoder}
Standard MAEs assume a fixed number of samples per patch, effectively tying the patch embedding to a particular sampling rate. 
Instead, we define patches \textit{in physical time} (i.e., in \textit{seconds}): each patch spans $\Delta=0.2$s, and therefore contains $L_r=r\Delta$ samples, where $r$ is the sampling rate.
Consequently, a 0.2s patch contains 2 samples at 10 Hz and 10 samples at 50 Hz.
Across sampling rates, fixing the patch duration in seconds enables consistency in the quantity of movement contained within a  patch.
Accelerometer data of window duration $d$ is split into $N = d/\Delta$ non-overlapping patches, and each patch $j$ covers $[j\Delta, (j{+}1)\Delta)$.

At a high-level, \ourmethod{} follows the standard MAE pipeline, but makes it input flexible.
A \textit{variable-rate tokenizer} first maps the $L_r$ samples in each visible patch to a single fixed size token vector.
A \textit{Transformer encoder} processes these tokens, and the temporal position is represented in physical seconds rather than indices.
The masked tokens are added back along with their positional embeddings, and a lightweight \textit{decoder} reconstructs the masked tokens through a \textit{variable-rate reconstructor head}, that produces $L_r$ samples at the required rate $r$.
In this way, the encoder-decoder remain identical across sampling rates as the input and output changes are handled by the tokenizer and reconstructor.
Following \cite{nie2022time}, we process \textit{each accelerometer channel independently}.

\textbf{1) Variable-Rate Tokenizer: }
As $L_r$ varies with sampling rate, the conventional linear patch projection cannot be applied.
Instead, we utilize a cross-attention tokenizer, similar to the Q-Former  \citep{li2023blip}. 
Each sample of accelerometer data is first projected to $d = 128$ and a single learned query vector then attends over $L_r$ embedded samples within the patch $\mathbf{E} = (\mathbf{e}_1, \dots, \mathbf{e}_{L_r})$ using 4 attention heads,
to produce a single patch token.
The learned query is input as the query whereas the key-value are the linear projections of the sample embeddings in the cross-attention setup.

The query vector is conditioned on the number of samples $L_r$ through an embedding of $\log L_r$ (a sinusoidal encoding followed by an MLP), thereby explicitly informing the tokenizer about the sampling rate.
In lieu of fixed sinusoidal encodings based on sample indices, the position of each sample within the patch is described as an offset in seconds: the sample $i$ sits at offset $\delta_i = (i{-}1)/r$ from the start of the patch.
At 50 Hz, a 10-sample patch has offsets $[0, 0.02, 0.04, \dots, 0.18]$ s; at 10 Hz, a 2-sample patch has offsets $[0, 0.1]$ s. 
Rotary Positional Embeddings (RoPE) are applied to the query and keys using these offset positions, and the learned query is positioned at the mean sample offset.
No tokenizer parameters depend on $L_r$, and it can process patches at different rates.
The tokenizer uses four attention heads, followed by a dense layer of size 256.

\textbf{2) Encoder-Decoder: }
The encoder ($f_{enc}$) is a four layer pre-LayerNorm (LN) Transformer with dim=128, four attention heads, and feedforward dim=256, followed by a LayerNorm (LN).
As in MAE \citep{he2022masked}, it only processes visible tokens. 
However, instead of indices, the tokens are positioned at the center $t_j$ of its patch (0.2s), which is independent of the sampling rate.

We first project the encoder outputs from 128$\rightarrow$64 dim, and inserts the masked tokens back at their original positions, to once again obtain $N$ patches.
Then, the self-decoder ($g_{self}$), a 2-layer pre-LN Transformer with dim=64, two attention heads, and feedforward dim=128 processes the $N$ patches.
The outputs at only the masked locations $\mathbf{h}_j$ are passed to the variable rate reconstruction head. 

\textit{Position Encoding:}
All attention layers utilize rotary position embeddings (RoPE) \citep{su2024roformer}.
The positions are specified in \textit{seconds} rather than token indices.
Hence, samples separated by one second have the same positional encoding regardless of the sampling rate (and as a result, the number of samples in between). 
For attention across patches, we use rotary periods spanning 0.04s to 120s, whereas within patch the maximum period is 1s.
This is done following Scale-MAE \citep{reed2023scale} which similarly conditions position encoding relative to the physical scale of the input. 
In addition, we employ QK-norm throughout \citep{henry2020query} as it has been shown to improve training stability \citep{dehghani2023scaling}.
As RoPE depends solely on relative position, the encoder is invariant to a global shift in timestamps.
This property enables us to separately encode the long and short views and subsequently align the embeddings for the cross resolution masked reconstruction.

\textbf{3) Variable-Rate Reconstruction: }
As a consequence of the variable rate input, the number of samples to be reconstructed also varies with the rate $r$ (0.2s at 10Hz is 2 samples but at 50Hz it is 10 samples). 
As a result, we replace the standard MAE's fixed reconstruction head with a small MLP conditioned on the patch's temporal offset (based on \citep{sitzmann2020implicit}).
For each masked patch, the transformer decoder produces a single representation $\mathbf{h}_j$, from which the MLP predicts the signal value for each temporal offset value in the patch. 
The predictions are conditioned on both the sampling rate and a sinusoidal encoding of the offset. 
Therefore, the MLP produces $r\Delta$ reconstructions at sampling rate $r$, allowing the same decoder to reconstruct signals at any sampling rate.
The variable rate decoders are shared for the masked reconstruction of each view, and a separate decoder is instantiated for the cross-resolution masked reconstruction.

\subsection{Cross-Resolution Masked Reconstruction}
\label{sec:method:objective}

The variable-rate MAE allows the same architecture to process different  sampling rates.
We exploit this capability during pre-training by constructing two views of the same accelerometer recording: (i) a long duration, high rate view $\mathbf{x}^{\mathrm{L}}$, and a shorter, potentially downsampled view $\mathbf{x}^{\mathrm{S}}$. 
Both views and encoded independently by the same encoder, with  $\mathbf{x}^{\mathrm{L}}$ providing additional temporal context through the cross-resolution masked reconstruction objective.

\textbf{1) Long and Short Views: }
$\mathbf{x}^{\mathrm{L}}$ has a fixed duration and sampling rate
$(d_{\mathrm{L}}, r_{\mathrm{L}}) = (30\mathrm{s}, 50\mathrm{Hz})$,
producing $N_{\mathrm{L}}=150$ patches. 
For each mini-batch, we randomly sample a short view duration $d_{\mathrm{S}} \in \{1,2,5,10\}\mathrm{s}$ and sampling rate $r_{\mathrm{S}} \in \{10,20,30,40,50\}\mathrm{Hz}$.
The entire mini-batch uses the same $(d_{\mathrm{S}}, r_{\mathrm{S}})$.
In $\mathbf{x}^{\mathrm{L}}$, we sample a start time $s$ on the patch grid and extract the $d_{\mathrm{S}}$-second crop, and then resample to $r_{\mathrm{S}}$ using polyphase resampling with an anti-aliasing filter to obtain $\mathbf{x}^{\mathrm{S}}$.
Here, cropping and downsampling create complementary information bottlenecks.
While cropping removes the surrounding temporal context, downsampling removes
fine-grained temporal information within $\mathbf{x}^{\mathrm{S}}$.

\textbf{2) Masking and Co-Masking: }
We first mask $\mathbf{x}^{\mathrm{S}}$ with a masking ratio of 0.75 using contiguous spans \citep{baevski2020wav2vec} of 5 tokens.
As $\mathbf{x}^{\mathrm{S}}$ is randomly cropped from $\mathbf{x}^{\mathrm{L}}$, the two views cannot be masked independently.
A region masked in $\mathbf{x}^{\mathrm{S}}$, if visible in $\mathbf{x}^{\mathrm{L}}$ would allow the cross-decoder ($g_{cross}$) to directly observe the reconstruction target.
Consequently, we perform masking on $\mathbf{x}^{\mathrm{S}}$ first, and match the masked locations in $\mathbf{x}^{\mathrm{L}}$ as well.
Every patch masked in $\mathbf{x}^{\mathrm{S}}$ is also masked at the corresponding location in $\mathbf{x}^{\mathrm{L}}$, and the patches visible in $\mathbf{x}^{\mathrm{S}}$ remain visible in $\mathbf{x}^{\mathrm{L}}$.
Additional patches in $\mathbf{x}^{\mathrm{L}}$ required to reach the 50\% masking ratio are sampled uniformly from outside the short-view interval.
This strategy of \textit{co-masking} removes the direct reconstruction shortcut while retaining the information advantage of $\mathbf{x}^{\mathrm{L}}$ due to the surrounding context. 

\textbf{3) Cross-Resolution Decoder: }
The cross-decoder ($g_{cross}$) reconstructs masked patches of $\mathbf{x}^{\mathrm{S}}$. 
Following \cite{gupta2023siamese}, the visible tokens from $\mathbf{x}^{\mathrm{L}}$ are used in the cross-attention setup as the key and value, whereas all tokens from $\mathbf{x}^{\mathrm{S}}$ (visible and masked) form the query.
The cross-decoder comprises two Transformer layers, with cross attention using $\mathbf{x}^{\mathrm{L}}$ and self-attention over $\mathbf{x}^{\mathrm{S}}$, and a feedforward block.
The reconstructed outputs at the masked locations of $\mathbf{x}^{\mathrm{S}}$ are converted back to raw signal values using a variable-rate reconstruction head.
Due to the co-masking, the cross-decoder must use the visible tokens of $\mathbf{x}^{\mathrm{S}}$ and the additional context from $\mathbf{x}^{\mathrm{L}}$ to reconstruct the masked tokens.

\textbf{4) Training Objective: }
We denote the visible tokens by $\mathcal{V}$, the masked patches by $\mathcal{M}$, and the encoder output by $\mathbf{Z}=f_\theta(\mathbf{x}_{\mathcal{V}})$, where $\mathbf{x}$ is the input signal. 
Let $\mathbf{z}^{S} = f_{\mathrm{enc}}(\mathbf{x}^{S}_{\mathcal{V}^{S}})$ and
$\mathbf{z}^{L} = f_{\mathrm{enc}}(\mathbf{x}^{L}_{\mathcal{V}^{L}})$ be the shared encoder's outputs on the visible patches of each view, with $\mathcal{M}^{S} \subseteq \mathcal{M}^{L}$ on the long view's time axis. $P^{S}$ and $P^{L}$ are the samples per patch in each view.
The overall training loss is the sum of the three masked reconstruction losses, given below:

\vspace{-1 em}
{\setlength{\jot}{1pt}
\begin{align}
\mathcal{L}_{\mathrm{Short}}
&= \frac{1}{|\mathcal{M}^{S}|P^{S}}
\sum_{t \in \mathcal{M}^{S}}
\big\| g_{\mathrm{self}}(\mathbf{z}^{S})_t - \mathbf{x}^{S}_t \big\|_2^2
\label{eqn:loss_short} \\
\mathcal{L}_{\mathrm{Long}}
&= \frac{1}{|\mathcal{M}^{L}|P^{L}}
\sum_{t \in \mathcal{M}^{L}}
\big\| g_{\mathrm{self}}(\mathbf{z}^{L})_t - \mathbf{x}^{L}_t \big\|_2^2
\label{eqn:loss_long} \\
\mathcal{L}_{\mathrm{Cross}}
&= \frac{1}{|\mathcal{M}^{S}|P^{S}}
\sum_{t \in \mathcal{M}^{S}}
\big\| g_{\mathrm{cross}}(\mathbf{z}^{S},\mathbf{z}^{L})_t - \mathbf{x}^{S}_t \big\|_2^2
\label{eqn:loss_cross}
\end{align}
}

\begin{equation}
\mathcal{L} = \mathcal{L}_{\mathrm{Cross}} + \mathcal{L}_{\mathrm{Short}} + \mathcal{L}_{\mathrm{Long}}
 \label{eqn:loss}
\end{equation}

The within-view losses $\mathcal{L}_{\mathrm{Short}}$ and $\mathcal{L}_ {\mathrm{Long}}$ reconstruct the masked samples of the short $\mathbf{x}^{\mathrm{S}}$ and long views $\mathbf{x}^{\mathrm{L}}$, respectively. 
$\mathcal{L}_{\mathrm{Short}}$ reconstructs the short view using only its own visible tokens, matching the downstream setting where no surrounding context is available.
The cross-resolution loss $\mathcal{L}_{cross}$ also reconstructs the masked tokens of $\mathbf{x}^{\mathrm{S}}$, but conditions the predictions on the \textit{visible tokens from the long view} $\mathbf{x}^{\mathrm{L}}$, through the cross-decoder.

\section{Experimental Settings}\label{sec:experimental_settings}

\textbf{Pre-training Data: }
We utilize data from the NHANES study \citep{cdc2010nhanes} for pre-training, similar to \cite{tarale2026bio}.
Specifically, we use all subjects from NHANES 2011--2012, NHANES 2013--2014, and NNYFS studies, leading to a total of $\sim$16k users, with approximately one week of data per user. 
The data were recorded at 80Hz and downsampled to 50Hz for pretraining.

\textbf{Evaluation Data: }
We evaluate \ourmethod{} on \textit{eight HAR datasets}, spanning fine-grained gestures, locomotion-style activities, and activities of daily living. 
The datasets vary substantially in sensor location, number of participants, and the number and types of classes, providing an effective testbed for evaluating transfer performance.
They include OpenWatch \citep{bonazzi2026openwatch} and WatchHand \citep{kim2026watchhand}, MobiAct \citep{chatzaki2016human} and MotionSense \citep{malekzadeh2021dana}, RealWorld \citep{sztyler2016body} and HARTH \citep{logacjov2021harth}, and Capture-24 \citep{chan2021capture} and PAAWS-SimFL \citep{potter2025physical}.
Full details in the Appendix (\cref{sec:app:eval_dataset_details}).

\subsection{Comparative Baselines}
We consider two evaluation settings for the baselines: \textit{(i)} controlled, where we pre-train each method on NHANES using our setup, thereby providing a direct comparison; 
and \textit{(ii)} public checkpoint, where we utilize publicly available pre-trained models to perform activity recognition on our downstream evaluation datasets. 
In this setting, all relevant data and method pre-processing remain identical to the original authors' implementations wherever applicable. 

For the controlled evaluation, we consider three representative SSL methods, with detailed information in the Appendix (\cref{sec:baseline_implementation_details}): 
\begin{enumerate}[label=(\arabic*), leftmargin=1.6em, itemsep=1pt, topsep=0pt, parsep=0pt]
    \item \textbf{Augmentation prediction} (AugPred; \citealp{saeed2019multi, yuan2024self}), where pre-training is performed via multi-task prediction of whether augmentations have been applied or not.
    \item \textbf{SimCLR} \citep{chen2020simple, tang2020exploring}, which involves contrastive learning between randomly augmented views of the same input window.
    \item \textbf{Vanilla MAE} \citep{he2022masked}, which uses an encoder-decoder identical to \ourmethod{}.
    We use the same patch duration (0.2s) at 50Hz, yielding a patch size of 10 samples.
    In contrast to \ourmethod{}, it uses linear projections for the patch embedding and for reconstruction. 
    This provides the most direct comparison to \ourmethod{}.
\end{enumerate}

For the public checkpoint evaluation, we use the public weights released by \cite{yuan2024self} (Harnet-Oxford), which were pre-trained on 700k person days of data from the UK Biobank.
We evaluate on both the 5s and 10s variants. 
Further, we also compare against BioPM \citep{tarale2026bio}, which is a recently released MAE pretrained on NHANES using 10s windows. 
Finally, we compare against general-purpose time-series FMs (TSFMs), including Chronos \citep{ansari2024chronos}, Chronos-2 \citep{ansari2025chronos} and MOMENT \citep{goswami2024moment}, to assess the benefit of specifically pretraining on accelerometer data. 
We embed each channel of accelerometer data separately and perform mean pooling for use in downstream activity recognition. 

\paragraph{Evaluation Setup}
In line with prior work \citep{saeed2019multi, haresamudram2022assessing, xu2025relcon}, we utilize subject-disjoint train/val/test splits.
For downstream datasets, we first randomly sample 20\% of users to form the test set. 
Of the remaining users, 20\% are once again sampled for validation and the rest is used for training. 
We repeat this process five times such that each user appears in the test set exactly once. 
For all results, we report the average macro-F1 score of the test set across folds. 
We provide all pretraining and evaluation parameters in the Appendix (\cref{sec:appendix_training_settings}). 

\textbf{Linear evaluation: } After pretraining, we freeze the encoder and train a single linear layer to predict activities.
The encoder functions purely as a feature extractor, allowing us to evaluate the quality of learned representations without confounding performance with classifier's capacity.
For all linear evaluations, we use Adam \citep{loshchilov2017decoupled} with lr=1e-3 and no weight decay.

\section{Results}

\subsection{Activity Recognition Performance}
\label{sec:activity_recognition_performance}
\begin{table}[t]
\centering
\caption{
    \textbf{Activity recognition performance:} we contrast \ourmethod{} against end-to-end training, general purpose TSFMs, as well as accelerometer-based FMs.
    \ourmethod{} matches or outperforms other FMs on six out of eight datasets. 
    \first{Bold} is best and \second{underline} second best per dataset.
    $^\dagger$ indicates settings where the window duration differs from that used during pretraining.
} 
\label{tab:main_results}
\vspace{2pt}
\footnotesize
\setlength{\tabcolsep}{2pt}
\renewcommand{\arraystretch}{1.15}
\setlength{\heavyrulewidth}{1.1pt}
\setlength{\lightrulewidth}{0.6pt}
\fitwidth{%
\begin{tabular}{@{}l >{\hspace{3pt}}r >{\hspace{5pt}}cc >{\hspace{5pt}}cc
                     >{\hspace{5pt}}cc >{\hspace{5pt}}cc@{}}
\toprule
\multirow{2}{*}{Approach}
  & \multirow{2}{*}{Enc \#Params}
  & \multicolumn{2}{c}{\ws{1}}
  & \multicolumn{2}{c}{\ws{2}}
  & \multicolumn{2}{c}{\ws{5}}
  & \multicolumn{2}{c}{\ws{10}} \\
\cmidrule(lr){3-4}\cmidrule(lr){5-6}\cmidrule(lr){7-8}\cmidrule(lr){9-10}
  &
  & \hd{Open\\Watch}  & \hd{Watch\\Hand}
  & \hd{Motion\\Sense} & \hd{Mobi\\Act}
  & \hd{HARTH}        & \hd{Real\\World}
  & \hd{PAAWS-\\SimFL} & \hd{Capture-\\24} \\
\midrule

\grp{\textit{End-to-end training}} \\ 
\meth{Conv encoder}
  & 80k & 0.5785 & \second{0.6846} & 0.8850 & 0.8125 & 0.7679 & 0.7420 & 0.7817 & 0.4177 \\
\meth{ResNet encoder}
  & 2.61M & 0.6398 & 0.6286 & 0.8775 & 0.8389 & 0.7933 & 0.7046 & \first{0.8364} & 0.4516 \\
\addlinespace[3pt]

\grp{\textit{TSFMs + linear eval}} \\
\meth{Chronos-T5-small}
  & 21M & 0.1182 & 0.4467 & 0.8303 & 0.5891 & 0.5876 & 0.6417 & 0.5894 & 0.4074 \\
\meth{Chronos-2}
  & 120M & 0.4466 & 0.6053 & 0.8163 & 0.7069 & 0.6530 & 0.7482 & 0.6782 & 0.4321 \\
\meth{MOMENT}
  & 385M & 0.3261 & 0.5572 & 0.7984 & 0.6071 & 0.6207 & 0.7305 & 0.6334 & 0.3893 \\
\addlinespace[3pt]

\grp{\textit{Public Checkpoints + linear eval}} \\
\meth{Harnet-Oxford}
  & 10M & 0.2030$^\dagger$ & 0.4991$^\dagger$ & 0.7494$^\dagger$ & 0.6426$^\dagger$ & 0.6412 & 0.7269 & 0.7820 & \first{0.5137} \\
\meth{BioPM}
  & 1.4M & 0.5133$^\dagger$
  & \first{0.7369}$^\dagger$
  & 0.8411$^\dagger$
  & 0.7000$^\dagger$
  & 0.7384$^\dagger$
  & 0.7526$^\dagger$
  & 0.6334 & 0.4039 \\
\addlinespace[3pt]

\grp{\textit{Controlled Specialist FMs + linear eval}} \\
\meth{AugPred}
  & 4.23M & 0.2666 & 0.4703 & 0.8704 & 0.7241 & 0.6732 & 0.7576 & 0.6465 & 0.4366 \\
\meth{SimCLR}
  & 4.23M & \first{0.6498} & 0.5473 & 0.8774 & 0.7633 & 0.6831 & 0.7574 & 0.6296 & 0.4344 \\
\meth{Vanilla MAE}
  & 0.53M & 0.5912 & 0.6513 & 0.8614 & 0.8313 & \second{0.8260} & 0.8068 & 0.7615 & 0.4519 \\
\addlinespace[3pt]

\grp{\textit{Ours}} \\
\meth{Ours (small)}
  & 0.70M & 0.6258 & 0.6675 & 0.8873 & 0.8441 & 0.8241 & 0.8179 & 0.7624 & 0.4450 \\
\meth{Ours (medium)}
  & 3.82M
  & \first{0.6491}
  & 0.6795
  & \first{0.9108}
  & \second{0.8643}
  & \first{0.8278}
  & \first{0.8299}
  & 0.7932
  & \second{0.4617} \\

\meth{Ours (large)}
  & 10.94M
  & \second{0.6457}
  & 0.6781
  & \second{0.9082}
  & \first{0.8664}
  & 0.8250
  & \second{0.8278}
  & \second{0.8002}
  & 0.4545 \\

\bottomrule

\end{tabular}%
}
\end{table}

\textbf{Public Checkpoint Evaluation: } 
In \cref{tab:main_results}, we compare \ourmethod{} against publicly released checkpoints from Harnet-Oxford \citep{yuan2024self} and BioPM \citep{tarale2026bio}.
Both methods were pretrained at 30Hz, we therefore resample all evaluation datasets to 30Hz before feature extraction. 
Scores marked with $^\dagger$ indicate settings where the downstream window duration differs from that used during pretraining. 
Harnet-Oxford provides two checkpoints pretrained using 5s and 10s windows.
We use the checkpoint whose pre-training window most closely matches the evaluation window.
For datasets with window sizes shorter than 5s, e.g., MobiAct, we perform circular padding to 5s before feature extraction to match the receptive field of the 5s Harnet-Oxford model.
We observe that this high quantity of padding (i.e., from 1s/2s to 5s) produces low performance.
The preprocessing for BioPM matches the original paper and requires no padding.

At 5s, \ourmethod{}-Small substantially outperforms Harnet-Oxford ($>$8\%), despite using a much smaller encoder (0.7M vs 10M params). 
On PAAWS-SimFL, Harnet-Oxford shows an increase over \ourmethod{}-Small (0.782 vs 0.762), while the Medium and Large models surpass it, achieving 0.793 and 0.800, respectively  (architecture details for all three model sizes are available in \cref{tab:model_sizes}).
All three variants of \ourmethod{} outperform BioPM on SimFL and Capture-24.
Interestingly, BioPM obtains the highest performance for WatchHand (73.69\%), despite substantial difference in window sizes.
This demonstrates that \ourmethod{} remains competitive against publicly released FMs, while scaling further improves performance. 

\textbf{Controlled Evaluation: }
As the released checkpoints provide matched comparisons only at the selected window durations, we additionally perform a controlled evaluation by pretraining specialist models for each of 1s, 2s, 5s, and 10s windows.
For AugPred and SimCLR, we use a shallower variant of the ResNet architecture from \cite{yuan2024self}, and adapt its receptive field to match the corresponding input duration (again, similar to \citealp{yuan2024self}). 
The Vanilla MAE uses the same encoder-decoder and patch duration as \ourmethod{}, making it our most direct comparison.
\ourmethod{}-Medium matches or \textit{outperforms all controlled baselines across datasets}.
Vanilla MAE is the strongest controlled baseline, highlighting the strength of masked reconstruction as a SSL pretext task, while also demonstrating the additional benefit of our flexible training approach.

\textbf{General-Purpose Time-Series FMs and End-to-End Training: }
All three variants of \ourmethod{} outperform the TSFMs indicating the substantial advantage of pretraining specifically using accelerometer data.
Despite freezing the encoder and only training a linear classifier, \ourmethod{} is highly competitive with end-to-end training, and
\ourmethod{}-Small outperforms both end-to-end models on four datasets, with the larger variants increasing it to six datasets. 

\subsection{Few-Shot Learning Evaluation}
\begin{figure}[t]
    \centering
    \includegraphics[width=0.85\linewidth]{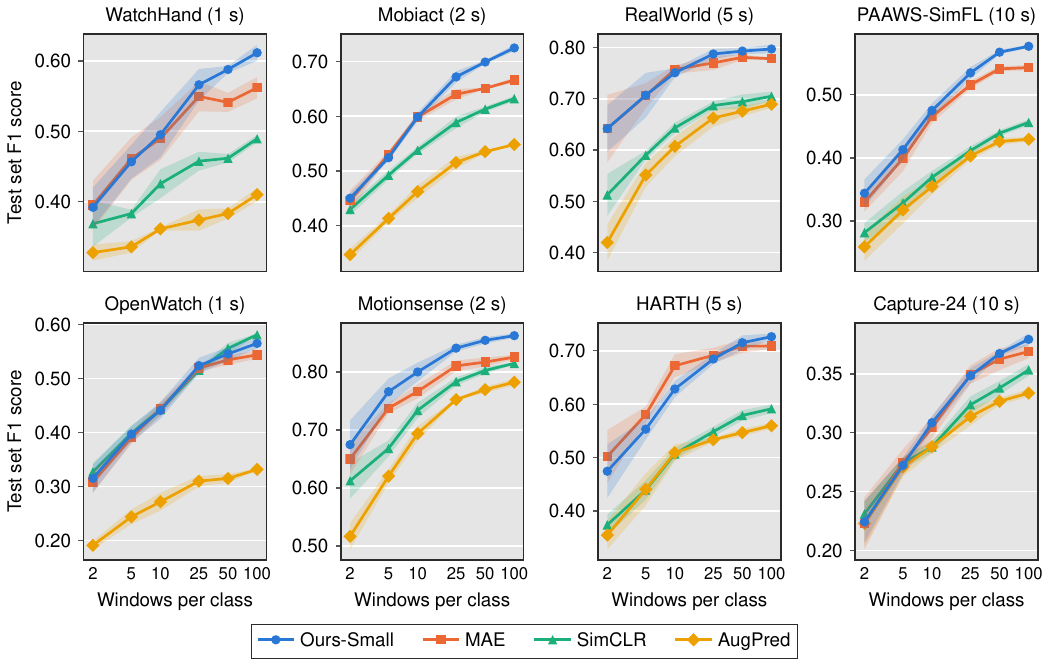}
    \caption{Few Shot Learning: During linear evaluation, we extract representations only for \{2, 5, 10, 25, 50, 100\} randomly sampled windows per activity, for five random seeds.
    Our approach matches or outperforms the specialized FM baselines.
    }
    \label{fig:fewshot}
\end{figure}
Annotating sensor data is a great challenge and is both costly and time consuming. 
Consequently, a key advantage of FMs is their ability to produce high performance when there is very limited access to labeled data, i.e., in the few-shot regime \citep{yuan2024self, haresamudram2022assessing}.  
In \cref{fig:fewshot} we contrast the few-shot learning performance of \ourmethod{}-Small against specialist FMs.

For each downstream dataset, we first randomly sample \{2, 5, 10, 25, 50, 100\} labeled windows per activity class from the train set, while validation and test sets remain untouched.
We extract representations using the pre-trained FMs and train a linear classifier using only the labeled windows. 
As performance can vary substantially in the limited data regime, we repeat the sampling and linear evaluation with five random seeds and report the test set performance across seeds in \cref{fig:fewshot}. 

Importantly, this setting is favorable to the specialist FMs as they are trained with the exact window duration and sampling rate required by the corresponding downstream datasets. 
In contrast, \ourmethod{} uses \textit{a single pre-trained model across datasets}.
As expected, the performance improves rapidly with the availability of increasing amounts of annotated data.
Overall, \ourmethod{} is competitive with or outperforms all specialist FMs, despite using a single shared model.
The vanilla MAE is the strongest specialist baseline and closest in performance in this few-shot regime.
As more labeled data are available, \ourmethod{} performs better, especially when the windows/activity $>$25.  
AugPred and SimCLR generally perform worse, often trailing by more than 10\% F1-score.

\subsection{Ablations}

\begin{figure}
    \centering
    \includegraphics[width=0.8\linewidth]{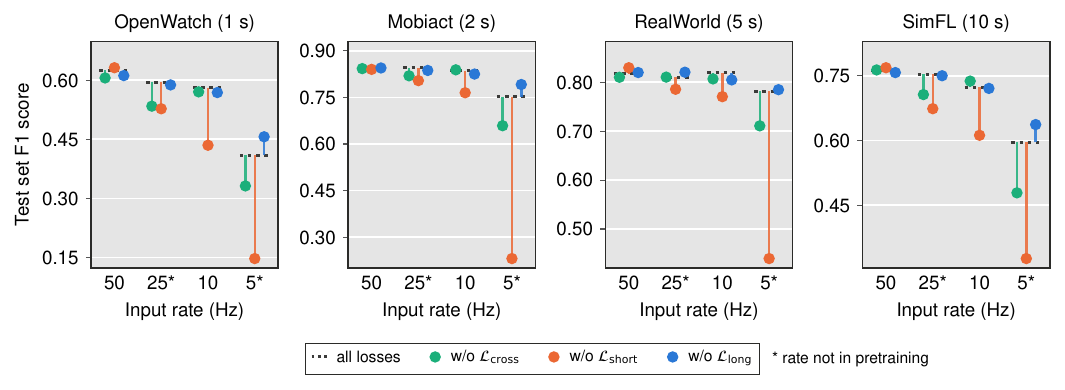}
    \caption{Impact of the pre-training losses: removing the short-view and cross-resolution losses has substantial impact at low frequencies.
}
    \label{fig:loss_ablation}
\end{figure}
\textbf{Contribution of the Losses:} We ablate each component of the pretraining objective (\cref{eqn:loss}) and evaluate the impact on four HAR datasets. 
We resample the datasets to \{5, 10, 25\}Hz in \cref{fig:loss_ablation} to assess the contribution of each component under changes in sampling rate. 
At 50Hz, removing any single loss has little effect on performance. 
The impact becomes substantially more pronounced as the sampling rate is progressively reduced.

The strongest effect is obtained with the removal of $\mathcal{L}_{\mathrm{Short}}$. 
At 5Hz, it produces substantial degradation in performance across datasets, with performance drops exceeding 25\% F1-score.
This is consistent with $\mathcal{L}_{\mathrm{short}}$ exposing the model to lower-rate signals during pretraining.
Without this objective, representations learned using only the higher rate view transfer poorly to heavily downsampled inputs.
The cross-resolution loss $\mathcal{L}_{\mathrm{cross}}$ is also important, and its removal produces a consistent reduction in performance. 
In contrast, $\mathcal{L}_{\mathrm{Long}}$ only has a modest impact across sampling rates.
As such, the utility of $\mathcal{L}_{\mathrm{short}}$ and $\mathcal{L}_{\mathrm{cross}}$ is primarily to retain performance as sampling rate decreases.

\textbf{Masking ratios: }
The default configuration of \ourmethod{} uses a 0.5 and 0.75 masking ratio for the long and short views, respectively.
In \cref{tab:mask_ratio_rates_mean}, we vary the masking ratios and report the mean of five-fold test set performance across WatchHand, MobiAct, RealWorld, and SimFL (full scores in the Appendix, \cref{tab:mask_ratio_rates_full}). 
At 50, 25, and 10 Hz, the performance is largely insensitive to the masking ratio in both views.
However, at the lowest rate (5 Hz), the masking ratio of the short-view has high impact.
Reducing from the default 0.75 to 0.5/0.25 causes performance drop of 14.5\% or higher.
In comparison, changing the long view's masking ratio from the default 0.5 to 0.25 or 0.75 has a substantially smaller effect at 5Hz, yielding 0.6709 and 0.7166 respectively.

These results mirror the findings in \cref{fig:loss_ablation}, where the design choices associated with the short view have little effect at higher sampling rates but become more important upon downsampling.
High masking ratios in the short-view encourage the models to retain their effectiveness even when less temporal resolution is available during evaluation.

\begin{table}[t]
\centering
\caption{
Varying the masking ratio during pre-training: we tabulate the mean test performance over four evaluation datasets, namely WatchHand, MobiAct, RealWorld, and SimFL. 
The short view's masking ratio has substantial impact at low sampling rates.
$^{*}$Rate not used in pretraining.
}
\label{tab:mask_ratio_rates_mean}
\fitwidth{
\begin{tabular}{cccccc|c}
\toprule
\multicolumn{2}{c}{Mask ratio} & & & & & \\
\cmidrule(lr){1-2}
Long view & Short view & 50\,Hz & 25\,Hz$^{*}$ & 10\,Hz & 5\,Hz$^{*}$ & Avg. \\
\midrule
0.5 & 0.75 & 0.7730 & 0.7681 & 0.7564 & 0.6874 & 0.7462 \\
\midrule
0.25 & 0.75 & 0.7739\ablup{0.0009} & 0.7707\ablup{0.0026} & 0.7565\ablup{0.0001} & 0.6709\abldn{0.0165} & 0.7430\abldn{0.0032} \\
0.75 & 0.75 & 0.7703\abldn{0.0027} & 0.7628\abldn{0.0053} & 0.7503\abldn{0.0061} & \textbf{0.7166}\ablup{0.0292} & 0.7500\ablup{0.0038} \\
0.5 & 0.25 & 0.7668\abldn{0.0062} & 0.7547\abldn{0.0134} & 0.7466\abldn{0.0098} & 0.5425\abldn{0.1449} & 0.7027\abldn{0.0435} \\
0.5 & 0.5 & \textbf{0.7751}\ablup{0.0021} & \textbf{0.7718}\ablup{0.0037} & \textbf{0.7616}\ablup{0.0052} & 0.4934\abldn{0.1940} & 0.7005\abldn{0.0457} \\
\bottomrule
\end{tabular}}
\end{table}


\section{Conclusion}
We introduced \ourmethod{}, a variable input MAE pretrained on large-scale accelerometry from $\sim$16k participants.
\ourmethod{} extends the standard MAE architecture with variable-rate tokenization and reconstruction, enabling a shared Transformer encoder-decoder to operate across different sampling rates and window durations.
During pretraining, we jointly optimized masked reconstruction within the long and short views, along with cross-resolution masked reconstruction, where the long-duration, high-frequency view provides context for reconstructing a shorter, downsampled view. 
We evaluated our approach on eight activity recognition datasets, covering a wide variety of activities, sensing locations, input window durations, and numbers of subjects.
These results demonstrate that a single accelerometer representation can remain effective across substantially different sensing configurations, reducing the need for bespoke training at specific input configurations.
Such flexibility is an important step towards the development of more broadly usable FMs for wearable sensing.


\subsection*{AI use statement}
In this work, we used generative AI tools to help write code, and to polish grammar and fix typos in the paper. 
We did not use them to develop ideas, to design the method or experiments, or to interpret results. 
The authors reviewed all AI-assisted work and take full responsibility for the contents of the paper.

\subsection*{Ethics statement}
Our work aims to support personalized health applications through developing models on physiological signals. We recognize that the use of such data raises important ethical considerations, particularly regarding privacy and potential health disparities, as increasingly detailed physiological characterization may expose sensitive information or disproportionately benefit certain populations. The collection, storage, and use of physiological data should therefore be subject to strict governance to ensure appropriate data protection and preserve individuals’ autonomy over their health information. In this study, all data are publicly available, and no personally identifiable information was used in our analyses. 
Overall, we believe that our work adheres to the ICLR Code of Ethics and contributes to more personalized health assessment, and, ultimately, improves the quality and safety of care.

\subsection*{Reproducibility statement}
To ensure complete reproducibility of our pre-training framework, model checkpoints and empirical findings, we will open-source all code, pretrained model weights, and downstream evaluation pipelines upon acceptance. 
\cref{sec:approach} describes the model architecture and the training objective. 
\cref{sec:experimental_settings} and the Appendix (\cref{sec:appendix_training_settings}) document the pretraining and downstream datasets, pre-processing procedures, data splits, optimization and training hyperparameters, downstream evaluation protocol, and implementation details for all controlled baselines. 
We additionally summarize the downstream datasets in the Appendix \ref{sec:appendix_datasets}.

\bibliography{refs}
\bibliographystyle{iclr2027_conference}

\appendix
\newpage
\section{Datasets}\label{sec:appendix_datasets}
\begin{table}[h]
  \centering
  \caption{
  Overview of the evaluated datasets, grouped by window length. 
  All signals are tri-axial accelerometry, downsampled to 50Hz (if their original frequencies are higher).
  The overlap is set to 50\% for the 2s and 5s datasets, 20\% for the 1s datasets, and 0 for Capture-24 and SimFL since they have a larger subject count.
  }
  \vspace{0.5em}
  \label{tab:datasets}
  \setlength{\tabcolsep}{5pt}
  \setlength{\dashlinedash}{0.6pt}
  \setlength{\dashlinegap}{1.6pt}
  \renewcommand{\arraystretch}{1.1}
  \small
  \begin{tabular}{@{}l>{\centering\arraybackslash}p{0.8cm}cc>{\raggedright\arraybackslash\footnotesize}p{6.2cm}@{}}
    \toprule
    Dataset & Win. & \#Users & \#Classes & \multicolumn{1}{l@{}}{Classes} \\
    \midrule
    OpenWatch   &               &  50 &  6 & double clench, double pinch, pinch down, pinch up, slide, negative \\
    WatchHand   & \winlabel{1}  &  23 &  3 & wrist up, wrist down, wrist left \\
    \midrule
    MobiAct     &               &  61 & 11 & standing, walking, jogging, jumping, stairs up, stairs down, stand-to-sit, sitting, sit-to-stand, car step-in, car step-out \\
    MotionSense & \winlabel{2}  &  24 &  6 & stairs down, stairs up, walking, jogging, standing, sitting \\
    \midrule
    RealWorld   &               &  15 &  8 & climbing down, climbing up, jumping, lying, running, sitting, standing, walking \\
    HARTH       & \winlabel{5}  &  22 &  9 & walking, running, shuffling, stairs up, stairs down, standing, sitting, lying, cycling \\
    \midrule
    Capture-24  &               & 151 & 10 & sleep, sitting, standing, walking, bicycling, vehicle, household chores, manual work, sports, mixed activity \\
    PAAWS-SimFL & \winlabel{10} & 250 &  9 & sitting, standing, lying, walking, stairs up, stairs down, gym exercises, biking, household chores \\
    \bottomrule
  \end{tabular}
\end{table}

\begin{figure}[h]
    \centering
    \includegraphics[width=1\linewidth]{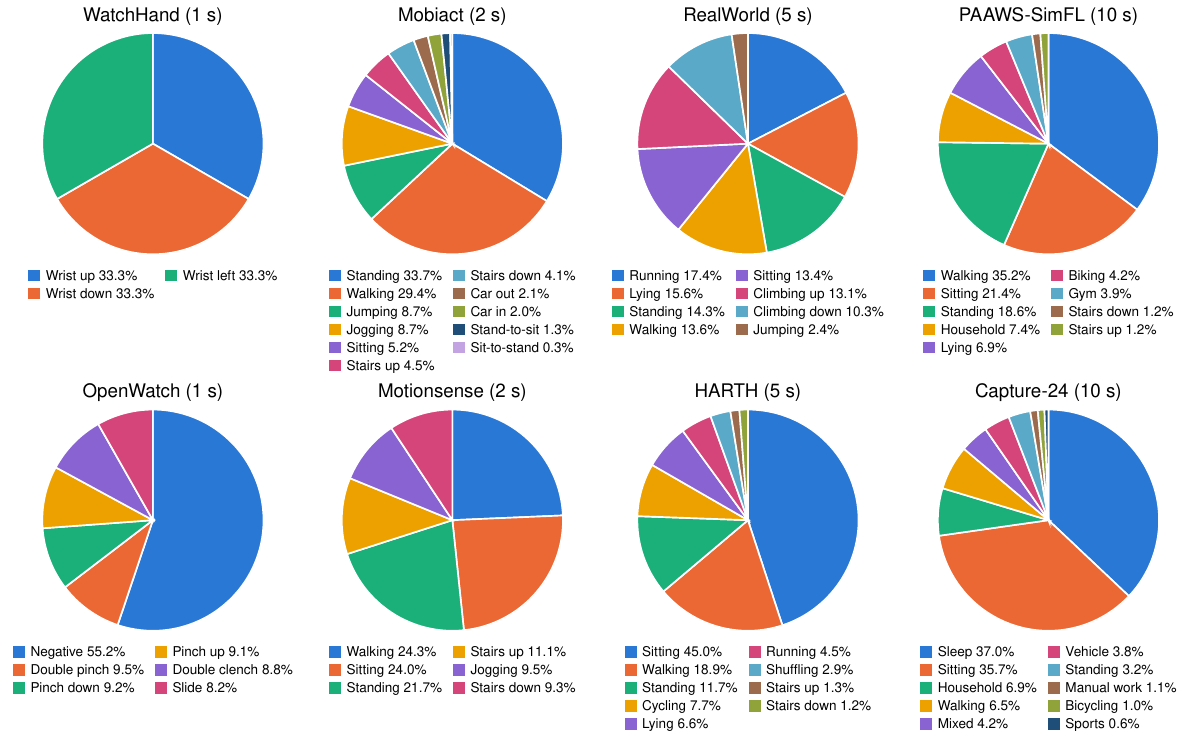}
    \caption{
    Visualizing the class proportions of the downstream datasets.
    Many datasets have substantial class imbalance.
    }
    \label{fig:class_proportions}
\end{figure}

\label{sec:app:eval_dataset_details}
We utilize the following window lengths for the datasets and provide a summary in \cref{tab:datasets}: 
\textit{(i)} OpenWatch \citep{bonazzi2026openwatch} and WatchHand \citep{kim2026watchhand} contain short wrist gestures such as pinches, clenches, and wrist rotations; following \cite{bonazzi2026openwatch}, we use a window size of 1s. 
\textit{(ii) }MobiAct \citep{chatzaki2016human} and MotionSense \citep{malekzadeh2021dana} contain locomotion-style activities recorded from smartphones at the waist/in the pocket, including sitting, walking, and jogging; following \cite{haresamudram2022assessing}, we use 2s windows. 
\textit{(iii) }RealWorld \citep{sztyler2016body} and HARTH \citep{logacjov2021harth} also comprise locomotion activities but have longer activity segments, motivating the use of longer, 5s windows, following \citet{liu2023focal, ustad2023har70}.
Finally, \textit{(iv) }Capture-24 \citep{chan2021capture} and PAAWS-SimFL \citep{potter2025physical} contain broader and more heterogeneous activities of daily living, such as housework, cycling, and sleeping, evaluated using 10s windows, based on prior work \citep{yuan2024self, potter2025physical}.
They are also the largest in terms of participant count.
Overall, these datasets span both small- and large-cohort settings with temporal contexts ranging from brief gestures to sustained and heterogeneous daily living activities.
This enables us to systematically evaluate if a single pre-trained model can perform effectively across substantially different settings. 

\section{Pre-training and Evaluation Settings}
\label{sec:appendix_training_settings}
For \ourmethod{} and all approaches in the controlled evaluation, we use the AdamW optimizer \citep{loshchilov2017decoupled} with a learning rate of $10^{-3}$ and no weight decay.
MAE and our method use $(\beta_1, \beta_2) = (0.9, 0.95)$, while SimCLR and AugPred use $(0.9, 0.999)$.
We pre-train for a maximum of 250k steps.
The learning rate is linearly warmed up over the first 20k steps and then decayed to zero using cosine-decay over the remaining steps.
Training uses bf16 mixed precision and distributed data parallelism across 2 GPUs, with a total batch size of 1024 (512 / GPU).

For NHANES, we randomly split users into training and validation at an 80:20 ratio. 
We define an epoch as 10k training batches, followed by evaluation using 1k validation batches.
We use early stopping with patience of five epochs, i.e., if the validation loss does not improve for five epochs, the training is terminated, and the checkpoint obtaining the lowest validation loss is used for downstream evaluation.

\paragraph{Batch Construction}
To construct the batch on each GPU, we first uniformly sample 64 training users without replacement.
Then, we sample one day per user and then draw eight 30\,s windows from that day, yielding $64\times8=512$ windows per GPU.
Windows are sampled with probability proportional to their standard deviation, so that stationary periods are sampled less often.
This sampling process is similar to the one used by \cite{yuan2024self}.

For the specialist FMs, the input window duration is one of \{1, 2, 5, 10\} seconds, which are obtained after cropping from the 30s segments.
The crop start time is also drawn with probability proportional to the standard deviation of the resulting crop.
The sampling rate is set to 50Hz.

For \ourmethod, each short-view has duration sampled randomly from $\{1, 2, 5, 10\}$~s and the input rate drawn uniformly from $\{10, 20, 30, 40, 50\}$~Hz.
The resulting short-view is paired with its corresponding full 30s window sampled at 50~Hz.

\begin{table}[!h]
\centering
\caption{Input configurations for accelerometer FMs.}
\label{tab:input_config_for_fms}
\small
\begin{tabular}{@{}lcc@{}}
\toprule
Foundation Model & Window (s) & Sampling rate (Hz) \\
\midrule
Harnet-Oxford \citep{yuan2024self} & 5/10 & 30 \\
BioPM \citep{tarale2026bio} & 10 & 30 \\
REBAR \citep{xu2024rebar} & 2.56 & 100 \\
RelCon \citep{xu2025relcon} & 2/2.56 & 100 \\
Oxford-Sleep FM \citep{yuan2024selfsleep} & 30 & 30 \\
\bottomrule
\end{tabular}
\end{table}

\subsection{Baseline Implementation Details}\label{sec:baseline_implementation_details}
The baselines in the controlled evaluation are trained within our experimental setup and therefore use identical data pre-processing and training settings (see details above).

\begin{enumerate}
    \item AugPred: We follow the setup of \cite{yuan2024self}, and utilize three augmentations provided in their implementation (\url{https://github.com/OxWearables/ssl-wearables}).
    The augmentations include: random permutation, time flipping, and time warping. 

    \item SimCLR: During pretraining, we randomly sample a pair of transforms for each batch from the following set: adding Gaussian noise, scaling, rotations, negation, time flipping, time warping, and channel shuffling. 
    The temperature for the InfoNCE loss is set to 0.1.
    The transforms are taken from: \url{https://github.com/iantangc/ContrastiveLearningHAR}.

    \item Vanilla MAE: this is the closest match to \ourmethod{} and uses a masking ratio of 0.5. It differs from our approach only in that it lacks the variable-rate tokenization and reconstruction heads.
    
\end{enumerate}

In the Public Checkpoint evaluation, we use Harnet-Oxford \citep{yuan2024self} and BioPM \citep{tarale2026bio}.
Harnet-Oxford was pretrained at 30Hz using window sizes of 5 or 10s, while BioPM was pre-trained using 10s windows also at 30Hz. 
We use code from the original repositories (\url{https://github.com/OxWearables/ssl-wearables} and \url{https://github.com/Prithvitarale/biopm}) to pre-process the evaluation datasets and extract representations according to the requirements of the public checkpoint.
For BioPM we additionally follow the original filtering and linear movement-element extraction pipeline.
The subsequent linear evaluation is performed within our setup, ensuring consistent evaluation protocol across methods. 

\subsection{Architecture Details}
\begin{table}[h]
\centering
\caption{Model size variants of \ourmethod{}. 
All sizes use 0.2\,s patches (10 samples at 50\,Hz) and a single-query tokenizer with 4 heads. 
}
\label{tab:model_sizes}
\fitwidth{%
\begin{tabular}{l cccc cccc cc}
\toprule
 & \multicolumn{4}{c}{Encoder} & \multicolumn{4}{c}{Decoder} & \multicolumn{2}{c}{Parameters} \\
\cmidrule(lr){2-5}\cmidrule(lr){6-9}\cmidrule(lr){10-11}
Model & Layers & Width & MLP & Heads & Layers & Width & MLP & Heads & Encoder & Total \\
\midrule
Small  & 4 & 128 & 256 & 4 & 2 & 64  & 128 & 2 & 0.70M  & 0.95M \\
Medium & 6 & 256 & 512 & 8 & 2 & 128 & 256 & 4 & 3.82M  & 4.81M \\
Large  & 8 & 384 & 768 & 8 & 2 & 192 & 384 & 4 & 10.94M & 13.17M \\
\bottomrule
\end{tabular}
}
\end{table}

\section{Results}
\begin{table}[h]
\centering
\caption{
Evaluating the impact of each loss component on performance. Bold: best within each dataset and rate. $^\dagger$Rate not used in pretraining.}
\label{tab:loss_ablation_rates}
\begin{tabular}{lcccc}
\toprule
 & 50\,Hz & 25\,Hz$^\dagger$ & 10\,Hz & 5\,Hz$^\dagger$ \\
\midrule
\multicolumn{5}{l}{\textit{OpenWatch (1\,s)}} \\
\quad All losses & 0.626 & \textbf{0.595} & \textbf{0.581} & 0.410 \\
\quad w/o $\mathcal{L}_{\mathrm{Cross}}$ & 0.606\abldn{0.020} & 0.534\abldn{0.061} & 0.571\abldn{0.010} & 0.332\abldn{0.078} \\
\quad w/o $\mathcal{L}_{\mathrm{Short}}$ & \textbf{0.632}\ablup{0.006} & 0.528\abldn{0.067} & 0.435\abldn{0.146} & 0.147\abldn{0.263} \\
\quad w/o $\mathcal{L}_{\mathrm{Long}}$ & 0.612\abldn{0.014} & 0.588\abldn{0.007} & 0.569\abldn{0.012} & \textbf{0.456}\ablup{0.046} \\
\midrule
\multicolumn{5}{l}{\textit{MobiAct (2\,s)}} \\
\quad All losses & 0.844 & \textbf{0.846} & 0.836 & 0.752 \\
\quad w/o $\mathcal{L}_{\mathrm{Cross}}$ & 0.843\abldn{0.001} & 0.820\abldn{0.026} & \textbf{0.839}\ablup{0.003} & 0.659\abldn{0.093} \\
\quad w/o $\mathcal{L}_{\mathrm{Short}}$ & 0.840\abldn{0.004} & 0.804\abldn{0.042} & 0.765\abldn{0.071} & 0.231\abldn{0.521} \\
\quad w/o $\mathcal{L}_{\mathrm{Long}}$ & \textbf{0.845}\ablup{0.001} & 0.837\abldn{0.009} & 0.825\abldn{0.011} & \textbf{0.792}\ablup{0.040} \\
\midrule
\multicolumn{5}{l}{\textit{RealWorld (5\,s)}} \\
\quad All losses & 0.818 & 0.810 & \textbf{0.819} & 0.782 \\
\quad w/o $\mathcal{L}_{\mathrm{Cross}}$ & 0.810\abldn{0.008} & 0.811\ablup{0.001} & 0.807\abldn{0.012} & 0.711\abldn{0.071} \\
\quad w/o $\mathcal{L}_{\mathrm{Short}}$ & \textbf{0.830}\ablup{0.012} & 0.786\abldn{0.024} & 0.771\abldn{0.048} & 0.440\abldn{0.342} \\
\quad w/o $\mathcal{L}_{\mathrm{Long}}$ & 0.820\ablup{0.002} & \textbf{0.821}\ablup{0.011} & 0.805\abldn{0.014} & \textbf{0.785}\ablup{0.003} \\
\midrule
\multicolumn{5}{l}{\textit{SimFL (10\,s)}} \\
\quad All losses & 0.762 & \textbf{0.753} & 0.723 & 0.596 \\
\quad w/o $\mathcal{L}_{\mathrm{Cross}}$ & 0.764\ablup{0.002} & 0.707\abldn{0.046} & \textbf{0.738}\ablup{0.015} & 0.479\abldn{0.117} \\
\quad w/o $\mathcal{L}_{\mathrm{Short}}$ & \textbf{0.769}\ablup{0.007} & 0.674\abldn{0.079} & 0.612\abldn{0.111} & 0.326\abldn{0.270} \\
\quad w/o $\mathcal{L}_{\mathrm{Long}}$ & 0.758\abldn{0.004} & 0.750\abldn{0.003} & 0.721\abldn{0.002} & \textbf{0.637}\ablup{0.041} \\
\bottomrule
\end{tabular}
\end{table}

\begin{table}[t]
\centering
\caption{
Evaluating the impact of the masking ratio of the long and short views. 
Bold: best within each dataset and rate. $^{*}$Rate not used in pretraining.}
\label{tab:mask_ratio_rates_full}
\begin{tabular}{lcccc}
\toprule
 & 50\,Hz & 25\,Hz$^{*}$ & 10\,Hz & 5\,Hz$^{*}$ \\
\midrule
\multicolumn{5}{l}{\textit{WatchHand (1\,s)}} \\
\quad default (long 0.5, short 0.75) & \textbf{0.668} & \textbf{0.664} & 0.647 & 0.620 \\
\quad long 0.25 & 0.667\abldn{0.001} & 0.661\abldn{0.003} & 0.644\abldn{0.003} & 0.609\abldn{0.011} \\
\quad long 0.75 & 0.667\abldn{0.001} & 0.653\abldn{0.011} & 0.642\abldn{0.005} & \textbf{0.633}\ablup{0.013} \\
\quad short 0.25 & 0.651\abldn{0.017} & 0.637\abldn{0.027} & 0.626\abldn{0.021} & 0.450\abldn{0.170} \\
\quad short 0.5 & 0.666\abldn{0.002} & 0.663\abldn{0.001} & \textbf{0.648}\ablup{0.001} & 0.381\abldn{0.239} \\
\midrule
\multicolumn{5}{l}{\textit{MobiAct (2\,s)}} \\
\quad default (long 0.5, short 0.75) & 0.844 & \textbf{0.846} & 0.836 & 0.752 \\
\quad long 0.25 & 0.843\abldn{0.001} & 0.841\abldn{0.005} & 0.839\ablup{0.003} & 0.744\abldn{0.008} \\
\quad long 0.75 & 0.840\abldn{0.004} & 0.835\abldn{0.011} & 0.830\abldn{0.006} & \textbf{0.790}\ablup{0.038} \\
\quad short 0.25 & 0.833\abldn{0.011} & 0.829\abldn{0.017} & 0.825\abldn{0.011} & 0.583\abldn{0.169} \\
\quad short 0.5 & \textbf{0.849}\ablup{0.005} & 0.843\abldn{0.003} & \textbf{0.842}\ablup{0.006} & 0.514\abldn{0.238} \\
\midrule
\multicolumn{5}{l}{\textit{RealWorld (5\,s)}} \\
\quad default (long 0.5, short 0.75) & 0.818 & 0.810 & \textbf{0.819} & 0.782 \\
\quad long 0.25 & 0.819\ablup{0.001} & 0.823\ablup{0.013} & 0.817\abldn{0.002} & 0.792\ablup{0.010} \\
\quad long 0.75 & 0.820\ablup{0.002} & 0.822\ablup{0.012} & 0.813\abldn{0.006} & \textbf{0.794}\ablup{0.012} \\
\quad short 0.25 & \textbf{0.831}\ablup{0.013} & 0.823\ablup{0.013} & 0.813\abldn{0.006} & 0.690\abldn{0.092} \\
\quad short 0.5 & 0.821\ablup{0.003} & \textbf{0.825}\ablup{0.015} & \textbf{0.819}\ablzero & 0.657\abldn{0.125} \\
\midrule
\multicolumn{5}{l}{\textit{SimFL (10\,s)}} \\
\quad default (long 0.5, short 0.75) & 0.762 & 0.753 & 0.723 & 0.596 \\
\quad long 0.25 & \textbf{0.766}\ablup{0.004} & \textbf{0.758}\ablup{0.005} & 0.727\ablup{0.004} & 0.538\abldn{0.058} \\
\quad long 0.75 & 0.754\abldn{0.008} & 0.741\abldn{0.012} & 0.716\abldn{0.007} & \textbf{0.649}\ablup{0.053} \\
\quad short 0.25 & 0.752\abldn{0.010} & 0.731\abldn{0.022} & 0.723\ablzero & 0.447\abldn{0.149} \\
\quad short 0.5 & 0.765\ablup{0.003} & 0.757\ablup{0.004} & \textbf{0.736}\ablup{0.013} & 0.422\abldn{0.174} \\
\bottomrule
\end{tabular}
\end{table}

\end{document}